\documentclass[10pt,twocolumn,letterpaper]{article}

\usepackage[pagenumbers]{cvpr} % To force page numbers, e.g. for an arXiv version

\usepackage{subcaption}  % For subfigures and subcaptions
\usepackage{amsmath}  % For advanced math
\usepackage{amssymb}  % Provides \mathbb
\usepackage{tabularray}
\usepackage{xcolor}

\definecolor{lightblue}{RGB}{220,235,255}

\definecolor{cvprblue}{rgb}{0.21,0.49,0.74}
\usepackage[pagebackref,breaklinks,colorlinks,allcolors=cvprblue]{hyperref}

\def\paperID{*****} % *** Enter the Paper ID here
\def\confName{CVPR}
\def\confYear{2026}

\title{U$^{2}$Flow: Uncertainty-Aware Unsupervised Optical Flow Estimation}

\author{
	Xunpei Sun$^{1}$,\quad
	Wenwei Lin$^{1}$,\quad
	Yi Chang$^{2}$,\quad
	Gang Chen$^{1}$\thanks{Corresponding author.}
	\and
	$^{1}$Sun Yat-sen University, Guangzhou, China\\
	$^{2}$Huazhong University of Science and Technology, Wuhan, China
	\vspace{-0.5em}
	\and
	{\tt\small sunxp7@mail2.sysu.edu.cn},\quad
	{\tt\small linww28@mail.sysu.edu.cn},\quad
	{\tt\small yichang@hust.edu.cn},\\
	{\tt\small cheng83@mail.sysu.edu.cn}
}

\begin{document}
\maketitle
\begin{abstract}
Unsupervised optical flow methods typically lack reliable uncertainty estimation, limiting their robustness and interpretability. We propose U$^{2}$Flow, the first recurrent unsupervised framework that jointly estimates optical flow and per-pixel uncertainty. 
The core innovation is a decoupled learning strategy that derives uncertainty supervision from augmentation consistency via a Laplace-based maximum likelihood objective, enabling stable training without ground truth. 
The predicted uncertainty is further integrated into the network to guide adaptive flow refinement and dynamically modulate the regional smoothness loss. 
Furthermore, we introduce an uncertainty-guided bidirectional flow fusion mechanism that enhances robustness in challenging regions. 
Extensive experiments on KITTI and Sintel demonstrate that U$^{2}$Flow achieves state-of-the-art performance among unsupervised methods while producing highly reliable uncertainty maps, validating the effectiveness of our joint estimation paradigm. 
The code is available at \url{https://github.com/sunzunyi/U2FLOW}.
\end{abstract}

\vspace{-0.5em}

\section{Introduction}
Optical flow estimation~\cite{sun2019models,dosovitskiy2015flownet,sunPWCNetCNNsOptical2018} is a fundamental vision task with broad applications~\cite{lin2022occlusionfusion,kim2022cross,zhong2024clearer,shen2023optical}.
Recently, deep recurrent models based on all-pairs correlation, such as RAFT~\cite{teedRAFTRecurrentAllPairs2021,sun2022disentangling,huang2022flowformer,shiFlowFormerMaskedCost2023,shiVideoFlowExploitingTemporal2023,luo2024flowdiffuser,dong2024memflow}, have achieved state-of-the-art performance in fully supervised settings.

However, obtaining large-scale, pixel-accurate ground-truth optical flow is costly and often impractical~\cite{sintel2012,sun2021autoflow,yuan2022optical,huang2023self}, motivating research on unsupervised and self-supervised optical flow estimation~\cite{liu2019ddflow,liu2019selflow,liuLearningAnalogyReliable2020}. Nevertheless, due to the absence of reliable supervision, self-supervised models often produce inaccurate estimates, especially when facing intrinsic challenges such as occlusions~\cite{wang2018occlusion,meister2018unflow,luoUPFlowUpsamplingPyramid2021}, textureless regions~\cite{jonschkowskiWhatMattersUnsupervised2020,marsal2023brightflow}, and large motion displacements~\cite{sun2025m2flow}. 
These estimation errors can be detrimental to downstream tasks. In point tracking~\cite{le2024dense} and multi-view reconstruction~\cite{poggi2020uncertainty}, even small local optical flow errors can accumulate and lead to failure. 
Moreover, erroneous flow estimates often cause artifacts in depth recovery~\cite{liu2020flow2stereo,zhou2025manydepth2} or motion-based segmentation~\cite{zhang2023optical}. 
Consequently, it is not enough for a model to predict what the motion is; it must also quantify how confident it is in that prediction~\cite{wannenwetsch2017probflow}. 

Despite its importance, uncertainty estimation in the self-supervised setting remains largely underexplored, primarily due to two core challenges: 
1) The Absence of Direct Supervision: Unlike supervised methods~\cite{ilg2018uncertainty,kang2022flownetu,shen2021realtime,iosif2023semanticflow}, which can be trained with ground-truth variance or likelihood information, self-supervised models lack any explicit “correct answer” for uncertainty. A fundamental difficulty lies in teaching a model to assess its own reliability without access to ground truth. 
2) The Effective Integration of Uncertainty: Even if uncertainty can be estimated, it is unclear how to leverage it effectively during training to improve flow accuracy, rather than treating it as a mere byproduct.

\begin{figure}[t]
	\centering
	\includegraphics{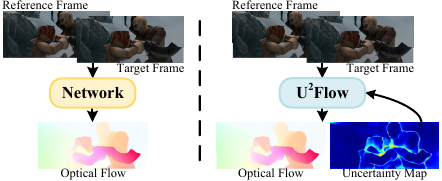}
	\caption{Comparison between previous optical flow estimation methods and our approach.
		(Left) Previous methods estimate only optical flow.
		(Right) Our proposed \textbf{U$^{2}$Flow} framework jointly estimates optical flow and its uncertainty, and further leverages the predicted uncertainty to refine the flow estimation.}
	\label{fig:our_output}
		\vspace{-1.5em}
\end{figure}

To address these challenges, we present \textbf{U$^{2}$Flow}, the first recurrent framework for the self-supervised joint estimation of dense optical flow and its per-pixel uncertainty, as illustrated in Fig.~\ref{fig:our_output}. 
First, to overcome the lack of direct uncertainty supervision, we derive a supervisory signal from the model's own predictive inconsistencies under data augmentation.  When the model yields inconsistent predictions under diverse spatial and appearance perturbations, it exposes its own regions of low confidence, thereby providing a powerful self-supervisory signal for uncertainty learning. By enforcing consistency between flows predicted under various perturbations, we derive a Laplace-based maximum likelihood objective~\cite{ilg2018uncertainty,gast2018probabilistic} that learns uncertainty distributions, decoupled from the main flow loss. 

Second, to effectively leverage the predicted uncertainty during training, we design an uncertainty-aware recurrent architecture. The predicted uncertainty is fed back into the recurrent update block to guide adaptive refinement. Concurrently, uncertainty is used to modulate the self-supervision objectives, enabling the model to intelligently down-weight unreliable signals.

Experiments show that U$^{2}$Flow achieves state-of-the-art performance among unsupervised methods on the KITTI and Sintel benchmarks. Crucially, it also produces highly reliable uncertainty maps,
demonstrating the efficacy of our joint estimation framework.

Our main contributions are summarized as follows:
\begin{itemize}
	\item To the best of our knowledge, we propose the first recurrent unsupervised framework for jointly estimating optical flow and uncertainty, which integrates an uncertainty prediction head and an uncertainty-aware refinement mechanism within the recurrent update block.

	\item We devise a decoupled uncertainty learning strategy based on augmentation consistency for stable estimation. Furthermore, we design an uncertainty-guided regional smoothness mechanism that leverages confidence to improve flow coherence.

	\item We propose an uncertainty-guided bidirectional flow fusion mechanism that utilizes uncertainty from both forward and backward flows to correct unreliable regions, outperforming traditional occlusion-based strategies.
\end{itemize}

\section{Related Work}

\noindent\textbf{Unsupervised Optical Flow:} 
Unsupervised optical flow estimation typically relies on photometric consistency losses combined with spatial smoothness regularization~\cite{ren2017unsupervised,meister2018unflow,jonschkowskiWhatMattersUnsupervised2020,luoUPFlowUpsamplingPyramid2021}. However, the photometric signal often becomes unreliable in the presence of occlusions, motion blur, illumination changes, or textureless regions~\cite{wang2018occlusion,liu2019ddflow,liu2019selflow,marsal2023brightflow,sun2025m2flow}. Early works~\cite{meister2018unflow,wang2018occlusion} explicitly handled occlusions by deriving binary masks to remove geometrically inconsistent pixels from the photometric loss.

Subsequent approaches, such as ARFlow~\cite{liuLearningAnalogyReliable2020} and SelFlow~\cite{liu2019selflow}, improved robustness through knowledge distillation and extensive data augmentation~\cite{liu2019ddflow,yuan2023semarflow,yuan2024unsamflow}. More recent efforts, including SemARFlow~\cite{yuan2023semarflow} and UnSAMFlow~\cite{yuan2024unsamflow}, incorporated semantic cues to enhance motion boundary preservation. Other methods~\cite{janaiUnsupervisedLearningMultiFrame2018,stone2021smurf,sun2025m2flow} leverage multi-frame training to provide richer temporal context and complementary motion evidence.

However, most models focus solely on point estimation and ignore prediction uncertainty, limiting their ability to differentiate between ambiguous and reliable regions.

\noindent\textbf{Uncertainty Estimation:}
Quantifying model confidence is crucial for building robust computer vision systems~\cite{wang2022itermvs,liu2024difflow3d,zhang2024diffsf,abdein2025self}. 
Early optical flow approaches typically treated uncertainty as a post-processing step~\cite{barron1994performance,kybic_cviu11,mac2012learning,kondermann2008statistical}, estimating confidence heuristically from image gradients~\cite{barron1994performance} or local flow energy~\cite{bruhn2006confidence} rather than integrating it into the learning process. 
Subsequent works introduced probabilistic formulations for joint estimation of optical flow and uncertainty~\cite{gal2016dropout,ilg2018uncertainty,yin2019hierarchical,gast2018probabilistic,wannenwetsch2017probflow}. 
ProbFlow~\cite{wannenwetsch2017probflow} uses variational inference within a probabilistic framework to jointly estimate flow and uncertainty. 
PDC-Net+~\cite{truong2023pdc} jointly learns dense correspondences and their associated uncertainties under supervised synthetic data, while 
ProbDiffFlow~\cite{zhou2025probdiffflow} outputs multiple flow hypotheses instead of a single result. 
Abdein et al.~\cite{abdein2025self} map the flow smoothness error to a probability distribution to model uncertainty.

Crucially, most existing optical flow uncertainty methods depend on full supervision~\cite{ilg2018uncertainty,kang2022flownetu,shen2021realtime,iosif2023semanticflow}, tightly coupling uncertainty learning with flow regression. 
This dependency makes them incompatible with the self-supervised paradigm where no such ground truth is available. 
Moreover, uncertainty is rarely leveraged to improve the flow estimation process itself during inference.

\section{Method}
\label{Method}

\begin{figure*}[htbp]
	\centering
	\subfloat[Overall network structure
	\label{fig:subfig1}]{
		\includegraphics{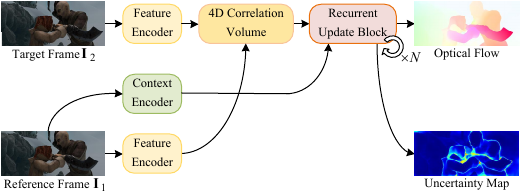}
	}
	\hspace{0.01\textwidth}
	\subfloat[Flow refinement module and uncertainty estimation head
	\label{fig:subfig2}]{
		\includegraphics{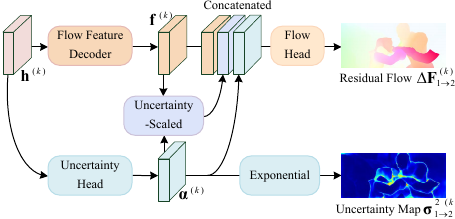}
	}
	\vspace{-0.5em}
    \caption{Overview of \textbf{U$^{2}$Flow} architecture. 
	(a) The overall recurrent structure follows RAFT~\cite{teedRAFTRecurrentAllPairs2021}. 
	(b) The uncertainty-aware refinement module and the uncertainty estimation head respectively predict optical flow and per-pixel uncertainty in the recurrent update block.}
	\label{fig:network-structure}
		\vspace{-1.25em}
\end{figure*}

Given two consecutive RGB frames 
\( \mathbf{I}_{1}, \mathbf{I}_{2} \in \mathbb{R}^{H \times W \times 3} \),  
our goal is to estimate the dense optical flow field 
\( \mathbf{F}_{1 \rightarrow 2} \in \mathbb{R}^{H \times W \times 2} \) 
along with its corresponding uncertainty map 
\( \boldsymbol{\sigma}^{2}_{1 \rightarrow 2} \in \mathbb{R}^{H \times W} \). 

\subsection{Network Overview}
\label{Method-1}

U$^{2}$Flow inherits the core design of RAFT~\cite{teedRAFTRecurrentAllPairs2021}, 
as illustrated in Fig.~\ref{fig:subfig1}. 
Given two consecutive frames \( \mathbf{I}_{1}, \mathbf{I}_{2} \), we first extract deep feature representations \( \mathbf{f}_{1}, \mathbf{f}_{2} \in \mathbb{R}^{H' \times W' \times D} \) and construct a 4D correlation volume $\mathbf{C} \in \mathbb{R}^{H' \times W' \times H' \times W'}$, 
which encodes pairwise similarities between all feature locations.  This correlation volume is iteratively queried by a recurrent update block that refines the optical flow estimate 
\( \mathbf{F}^{(k)}_{1 \rightarrow 2} \) through multiple iterations \( k = 1, 2, \ldots, K \).

Following SMURF~\cite{stone2021smurf}, we replace all batch normalization layers with instance normalization to enhance training stability and convergence under small-batch, unsupervised settings. 
Furthermore, U$^{2}$Flow introduces an additional uncertainty head to estimate the flow uncertainty \( \boldsymbol{\sigma}^{2\,(k)}_{1 \rightarrow 2} \) at each iteration. 
This naturally raises an important question: can the estimated uncertainty be utilized within the network itself to further refine the optical flow? 

To address this, we reformulate the original flow head into an uncertainty-aware refinement module that leverages the predicted uncertainty to guide the iterative flow refinement process (see Sec.~\ref{Method-2}). 
Meanwhile, during training, the refined flow supervises the uncertainty branch, progressively enhancing its estimation accuracy (see Sec.~\ref{Method-3}). 
In this way, U$^{2}$Flow jointly learns optical flow and uncertainty within a unified optimization framework.

\subsection{Uncertainty Estimation and Flow Refinement}
\label{Method-2}
As depicted in Fig.~\ref{fig:subfig2}, after obtaining the hidden representation from the RAFT recurrent unit, U$^{2}$Flow first applies an uncertainty estimation head and then feeds its output into our refinement module. Given the hidden feature representation \( \mathbf{h}^{(k)} \) at iteration \( k \), we first decode an intermediate flow feature and estimate its corresponding uncertainty as 
\begin{equation}
	\mathbf{f}^{(k)} = \mathcal{C}_{\text{flow}}(\mathbf{h}^{(k)}), 
	\qquad 
	\boldsymbol{\alpha}^{(k)} = \mathcal{C}_{\text{unc}}(\mathbf{h}^{(k)}),
\end{equation}
where \( \mathcal{C}_{\text{flow}}(\cdot) \) and \( \mathcal{C}_{\text{unc}}(\cdot) \) 
denote convolutional layers for flow feature extraction and uncertainty estimation, respectively. 
To ensure strictly positive uncertainty values, the uncertainty head predicts the logarithm of the flow variance:
\begin{equation}
	\boldsymbol{\alpha}^{(k)} = \log \left( \boldsymbol{\sigma}^{2\,(k)}_{1 \rightarrow 2} \right),
\end{equation}
which improves numerical stability during training. During inference, the actual flow uncertainty (i.e., the variance) is recovered as 
\( \boldsymbol{\sigma}^{2\,(k)}_{1 \rightarrow 2} = \exp(\boldsymbol{\alpha}^{(k)}) \).

To modulate the influence of uncertain regions, an uncertainty weight map is obtained via a sigmoid transformation:
\begin{equation}
	\mathbf{s}^{(k)} = \phi(-\boldsymbol{\alpha}^{(k)}),
\end{equation}
which acts as a reliability indicator for each flow vector. The uncertainty-scaled flow feature is defined as
\begin{equation}
	\tilde{\mathbf{f}}^{(k)} = \mathbf{f}^{(k)} \odot \mathbf{s}^{(k)*},
\end{equation}
where \( \odot \) denotes element-wise multiplication, and \( (\cdot)^{*} \) represents a stop-gradient operation that prevents backpropagation through the uncertainty branch.

Finally, the refined optical flow residual is obtained by fusing the original feature, the scaled feature, and the uncertainty map:
\begin{equation}
	\Delta \mathbf{F}^{(k)}_{1 \rightarrow 2} = 
	\mathcal{C}'_{\text{flow}}\!\left(
	\text{concat}\!\left(\mathbf{f}^{(k)},\, \tilde{\mathbf{f}}^{(k)},\, \boldsymbol{\alpha}^{(k)*}\right)
	\right),
\end{equation}
where \( \mathcal{C}'_{\text{flow}}(\cdot) \) denotes the convolutional head that outputs the flow residual \( \Delta \mathbf{F}^{(k)}_{1 \rightarrow 2} \in \mathbb{R}^{H' \times W' \times 2} \).  

This uncertainty-aware refinement mechanism enables the model to dynamically use predicted uncertainty to modulate flow features, effectively suppressing the influence of unreliable regions during refinement.

\subsection{Uncertainty-Aware Unsupervised Loss}\label{Method-3}
Unlike standard unsupervised optical flow methods, our loss explicitly integrates predicted uncertainty to modulate the training signal, enabling joint flow and uncertainty estimation within a purely self-supervised framework.

\noindent\textbf{Photometric Loss:}~~
During each refinement iteration \( k \), the input frames are warped by \( \mathbf{F}_{1 \rightarrow 2}^{(k)} \) and \( \mathbf{F}_{2 \rightarrow 1}^{(k)} \) to synthesize a new view of the source frame. 
Similar to ARFlow~\cite{liuLearningAnalogyReliable2020}, the photometric loss \( \ell_{\mathrm{ph}}^{(k)} \) between each original image and its warped counterpart is computed as a weighted combination of three terms: 
the pixel-wise \(\ell_1\) distance, SSIM, and the census loss~\cite{meister2018unflow}. 
The occlusion mask $\mathbf{O}_{i \rightarrow j}^{(k)}$ is computed using a forward--backward consistency check~\cite{meister2018unflow} to exclude regions without valid correspondences:
\begin{equation}
	\mathbf{O}_{i \rightarrow j}^{(k)}(p)
	= \mathbb{1}\Bigl(\bigl\|\mathbf{F}_{i \rightarrow j}^{(k)}(p) +
	\mathbf{F}_{j \rightarrow i}^{(k)}(p + \mathbf{F}_{i \rightarrow j}^{(k)}(p))\bigr\|_{2} 
	> \delta \Bigr),
	\label{eq:fbc}
\end{equation}
where $\delta$ represents an adaptive threshold, $\mathbb{1}(\cdot)$ is the indicator function, $(i,j) \in \{(1,2), (2,1)\}$ refers to the flow direction, and $p$ indicates the pixel coordinates.

\noindent\textbf{Smoothness Loss:}\label{hm_loss} 
To encourage locally coherent flow while preserving motion boundaries, we apply an edge-aware smoothness regularization to the predicted flow $\mathbf{F}_{i \rightarrow j}^{(k)}$ at each iteration $k$, denoted as $\ell_{\mathrm{sm}}^{(k)}$. Building on the work of UnSAMFlow~\cite{yuan2024unsamflow}, we also incorporate a regional smoothness constraint based on homography estimation. For each object region, a homography is estimated via RANSAC from reliable correspondences derived from the current flow prediction. 
The resulting homographies are then used to generate regionally refined flow fields.

\begin{figure}[t]
	\centering
	\includegraphics{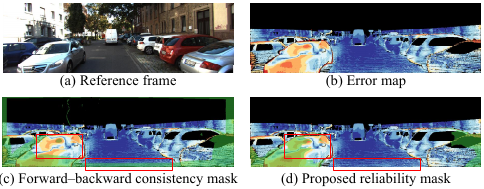}
	\caption{Comparison of different masks for indicating flow errors. 
		The masks are visualized as translucent \textcolor{green}{green} overlays on the optical flow error maps, 
		where correct estimations are shown in \textcolor{blue}{blue}, incorrect ones in \textcolor{red}{red}, 
		and black pixels denote regions without ground truth. 
		As shown in (c), some high-error regions remain unmasked, while certain low-error regions are incorrectly masked. 
		In contrast, our method (d) accurately identifies high-error regions, providing more reliable cues for 
		subsequent homography smoothness (Sec.~\ref{hm_loss}), flow fusion (Sec.~\ref{Method-4}), and other downstream tasks.
		}
	\label{fig:fusion_mask}
		\vspace{-1.0em}
\end{figure}

A key distinction of our approach lies in how reliable correspondences are identified. While the original formulation employs occlusion masks to exclude unreliable pixels, we introduce a more principled uncertainty-based reliability mask. This mechanism leverages the model's predicted uncertainty, excluding pixels with uncertainty above a threshold $\tau_{\text{hg}}$ (see Fig.~\ref{fig:fusion_mask}). It ensures that only high-confidence regions contribute to homography estimation and optimization. Consequently, the smoothness regularization becomes adaptive, directly guided by the model's learned confidence. It is worth noting that due to the stringent planarity assumption of homography, this uncertainty-enhanced component is applied exclusively to the KITTI~\cite{geiger2013vision, menze2015object} dataset, which contains predominantly planar rigid motions. The effectiveness of this uncertainty-guided strategy is further validated in our ablation studies (Sec.~\ref{E-A-1}).

The homography smoothness loss is defined as the $\ell_{1}$ distance between the predicted flow $\mathbf{F}_{i \rightarrow j}$ and the homography-refined flow $\mathbf{F}^{\mathrm{H}}_{i \rightarrow j}$:
\begin{equation}
	\ell_{\mathrm{hg}} = 
	\frac{1}{H' \times W'}
	\sum\nolimits_{\boldsymbol{p}}
	\left\|
	\mathbf{F}_{i \rightarrow j}(\boldsymbol{p}) -
	\mathbf{F}^{\mathrm{H}}_{i \rightarrow j}(\boldsymbol{p})
	\right\|_{1}.
\end{equation}

\noindent\textbf{Augmentation and Uncertainty Supervision:}
In the absence of ground-truth, we generate a supervisory signal for the uncertainty head through the principle of augmentation consistency. The process begins by computing an initial flow estimate, $\mathbf{F}_{1 \rightarrow 2}$, for an image pair $(\mathbf{I}_{1}, \mathbf{I}_{2})$ in an initial forward pass. We then apply a set of strong appearance and spatial augmentations to both the images and the flow field, producing an augmented pair $(\hat{\mathbf{I}}_{1}, \hat{\mathbf{I}}_{2})$ and a transformed pseudo-ground-truth flow $\hat{\mathbf{F}}_{1 \rightarrow 2}$. The network then re-estimates the flow on the augmented pair, yielding a new prediction $\hat{\mathbf{F}}^{\prime}_{1 \rightarrow 2}$. The $\ell_{1}$ distance between these flows, $\hat{D}^{(k)}(\boldsymbol{p}) = \|\hat{\mathbf{F}}_{1 \rightarrow 2}(\boldsymbol{p}) - \hat{\mathbf{F}}^{\prime(k)}_{1 \rightarrow 2}(\boldsymbol{p})\|_1$, serves as the self-supervised target, as it captures the model's predictive inconsistency under perturbation.

This augmentation consistency introduces diverse appearance perturbations (e.g., color jitter, contrast adjustment, Gaussian noise, random erasing) and spatial transformations (e.g., translation, random rotation, rescaling), thereby exposing the network to a wide range of uncertainty conditions. 
To capture this uncertainty, we adopt a Maximum Likelihood Estimation (MLE) formulation.  
For each augmentation iteration $k$, the objective is to minimize the negative log-likelihood (NLL) of observing the target flow $\hat{\mathbf{F}}_{1 \rightarrow 2}$ given the predicted distribution:
\begin{equation}
	\mathcal{L}_{\text{unc}}^{(k)} = 
	- \mathbb{E}_{\boldsymbol{p}}\!\left[
	\log p\!\left(\hat{\mathbf{F}}_{1 \rightarrow 2}(\boldsymbol{p}) 
	\mid 
	\hat{\mathbf{F}}^{\prime(k)}_{1 \rightarrow 2}(\boldsymbol{p}), 
	\boldsymbol{\sigma}^{2(k)}(\boldsymbol{p})\right)
	\right].
	\label{eq:nll_general}
\end{equation}

Following~\cite{ilg2018uncertainty}, we assume a Laplace likelihood, which aligns naturally with the $\ell_{1}$-based residual $\hat{D}^{(k)}$.  
Under the independence assumption across flow dimensions, this leads to the numerically stable MLE objective:
\begin{equation}
	\begin{cases}
		\tilde{\ell}_{\mathrm{unc}}^{(k)}(\boldsymbol{p}) =
		\sqrt{2}\, \exp\!\left(-\tfrac{1}{2} \boldsymbol{\alpha}^{(k)}(\boldsymbol{p})\right)
		\hat{D}^{(k)}(\boldsymbol{p})
		+ \tfrac{1}{2} \boldsymbol{\alpha}^{(k)}(\boldsymbol{p}), \\[8pt]
		\ell_{\mathrm{unc}}^{(k)} =
		\dfrac{
			\sum_{\boldsymbol{p}} 
			(1 - \hat{O}_{1 \rightarrow 2}(\boldsymbol{p})) \,
			\tilde{\ell}_{\mathrm{unc}}^{(k)}(\boldsymbol{p})
		}{
			\sum_{\boldsymbol{p}} 
			(1 - \hat{O}_{1 \rightarrow 2}(\boldsymbol{p}))
		}.
	\end{cases}
	\label{eq:uncloss}
\end{equation}
Here, $\boldsymbol{\alpha}^{(k)} = \log \boldsymbol{\sigma}^{2(k)}$ denotes the predicted log-variance, and $\hat{O}_{1 \rightarrow 2}$ is the transformed occlusion map~\cite{liuLearningAnalogyReliable2020}.
Unlike prior works~\cite{ilg2018uncertainty,kang2022flownetu,shen2021realtime,iosif2023semanticflow} that tightly couple flow regression and uncertainty estimation via a single MLE objective, our framework adopts a decoupled design, separating the flow loss from the uncertainty loss. 
Consequently, we detach $\hat{D}^{(k)}$ from the MLE objective to prevent gradient leakage into the main flow estimation branch, which improves the robustness and stability of self-supervised learning (see Sec.~\ref{E-A-1} for ablations).

\begin{table*}[htbp]
	\centering
	\fontsize{9pt}{11pt}\selectfont
	\begin{tblr}{
			colspec={c|l|cc|cc|c ccc|c cc}, 
			stretch=0.5
		}
		\hline[1.25pt]
		\SetCell[r=3, c=1]{c} {}
		& \SetCell[r=3, c=1]{l} {Method} 
		& \SetCell[c=2]{c} {Sintel Clean} & 
		& \SetCell[c=2]{c} {Sintel Final} & 
		& \SetCell[c=4]{c} {KITTI 2012} & & & 
		& \SetCell[c=3]{c} {KITTI 2015} & & \\
		\cline{3-13}
		& & train & test & train & test & train & \SetCell[c=3]{c} {test} & & & train & \SetCell[c=2]{c} {test} & \\
		\cline{3-13}
		& & EPE & EPE & EPE & EPE & EPE & Fl-all & Fl-noc & EPE & EPE & Fl-all & Fl-noc \\
		\hline
		\SetCell[r=5, c=1]{c} \rotatebox{90}{Supervised} 
		& PWC-Net+ \cite{sun2019models} 
		& (1.71) & 3.45 & (2.34) & 4.60 & (0.99) & 6.72 & 3.36 & 1.4 & (1.47) & 7.72 & 4.91 \\
		& RAFT \cite{teedRAFTRecurrentAllPairs2021} 
		& (0.77) & 1.61 & (1.27) & 2.86 & -- & -- & -- & -- & (0.63) & 5.10 & 3.07 \\
		& FlowFormer \cite{huang2022flowformer}
		& (0.48) & 1.16 & (0.74) & 2.09 & -- & -- & -- & -- & (0.53) & 4.68 & 2.69 \\
		
		& VideoFlow \cite{shiVideoFlowExploitingTemporal2023}$^{MF}$ 
		& (0.46) & \textbf{0.99} & (0.66) & \textbf{1.62} & -- & -- & -- & -- & (0.56) & \textbf{3.65} & -- \\
		& FlowDiffuser \cite{luo2024flowdiffuser}
		& -- & 1.02 & -- & 2.03 & -- & -- & -- & -- & -- & 4.17 & 2.82 \\
		\hline
		\SetCell[r=10, c=1]{c} \rotatebox{90}{Unsupervised}
		& UnFlow-CSS \cite{meister2018unflow}
		& -- & 9.38 & (7.91) & 10.22 & 3.29 & -- & -- & -- & 8.10 & 23.27 & -- \\
		& SelFlow \cite{liu2019selflow}$^{MF}$ 
		& (2.88) & 6.56 & (3.87) & 6.57 & 1.69 & 7.68 & 4.31 & 2.2 & 4.84 & 14.19 & 9.65 \\
		& UFlow \cite{jonschkowskiWhatMattersUnsupervised2020} 
		& (2.50) & 5.21 & (3.39) & 6.50 & 1.68 & 7.91 & 4.26 & 1.9 & 2.71 & 11.13 & 8.41 \\
		& ARFlow \cite{liuLearningAnalogyReliable2020} 
		& (2.79) & 4.78 & (3.73) & 5.89 & 1.44 & -- & -- & 1.8 & 2.85 & 11.80 & -- \\
		& UPFlow \cite{luoUPFlowUpsamplingPyramid2021} 
		& (2.33) & 4.68 & (2.67) & 5.32 & 1.27 & -- & -- & 1.4 & 2.45 & 9.38 & -- \\
		& SMURF \cite{stone2021smurf}$^{MF}$ 
		& (1.71) & 3.15 & (2.58) & 4.18 & -- & \textbf{6.19} & \textbf{3.13} & 1.4 & 2.00 & 6.83 & 5.26 \\
		& SemARFlow \cite{yuan2023semarflow}$^{\dagger}$ 
		& -- & -- & -- & -- & 1.28 & 7.35 & 3.90 & 1.5 & 2.18 & 8.38 & 5.43 \\
		& UnSAMFlow \cite{yuan2024unsamflow}$^{\dagger}$ 
		& (2.21) & 3.93 & (3.07) & 5.20 & 1.26 & 7.05 & 3.79 & 1.4 & 2.01 & 7.83 & 5.67 \\
		& M2Flow \cite{sun2025m2flow}$^{MF}$ 
		& (2.01) & 3.38 & (3.12) & 5.01 & \textbf{1.09} & \underline{6.24} & 3.95 & \textbf{1.2} & 1.95 & 7.37 & 5.73 \\
		& U$^2$Flow (Ours)
		& (\underline{1.42}) & \underline{2.83} & (\underline{2.32}) & \underline{4.16} & 1.19 & 6.37 & 3.48 & 1.4 & \underline{1.83} & \underline{6.13} & \underline{4.56} \\
		& U$^2$Flow (Ours +FF)$^{MF}$ 
		& (\textbf{1.36}) & \textbf{2.83} & (\textbf{2.29}) & \textbf{4.10} & \underline{1.12} & 6.26 & \underline{3.47} & \underline{1.3} & \textbf{1.74} & \textbf{6.00} & \textbf{4.52} \\
		\hline[1.25pt]
	\end{tblr}	
	\caption{Quantitative results on Sintel and KITTI online benchmarks. Metrics evaluated at ``all" (all pixels), ``noc" (non-occlusions). \textit{MF} denotes methods trained using multi-frame data. ${\dagger}$ denotes models with semantic inputs. ``+FF" denotes our bidirectional flow fusion module. Missing entries (--) denote unreported results. Parentheses indicate that training and testing are conducted on the same dataset.}
	\label{tab:merged_benchmarks}
		\vspace{-1.0em}
\end{table*}

To ensure the flow estimation branch continues to benefit from augmentation regularization, we retain the standard augmentation loss $\ell_{\mathrm{ar}}^{(k)}$, which operates on the same residual $\hat{D}^{(k)}$ with gradient flow enabled, and further incorporate semantic augmentations~\cite{yuan2024unsamflow}.
The augmentation regularization loss is defined as the $\ell_1$ distance between the transformed and predicted flows:
\begin{equation}
	\ell_{\mathrm{ar}}^{(k)} = 
	\dfrac{
		\sum_{\boldsymbol{p}} 
		(1 - \hat{O}_{1 \rightarrow 2}(\boldsymbol{p})) 
		\hat{D}^{(k)}(\boldsymbol{p})
	}{
		\sum_{\boldsymbol{p}} 
		(1 - \hat{O}_{1 \rightarrow 2}(\boldsymbol{p}))
	}.
	\label{eq:ar_aug}
\end{equation}

A similar formulation applies for semantic augmentations, denoted as $\ell_{\mathrm{sem}}^{(k)}$~\cite{yuan2024unsamflow}.

\noindent\textbf{Final Loss:}  
The overall training objective integrates all loss components as follows:
{\small
\begin{equation}
	\label{eq:tloss}
	\begin{split}
		\ell_{\text{Total}} 
		&= \lambda_{\text{hg}} \, \ell_{\mathrm{hg}} 
		+ \sum_{k=1}^{K} \zeta^{K-k} \Big( 
		\ell_{\mathrm{ph}}^{(k)} 
		+ \lambda_{\text{sm}} \, \ell_{\mathrm{sm}}^{(k)} 
		+ \lambda_{\text{ar}} \, \ell_{\mathrm{ar}}^{(k)}  \\ 
		&\quad\quad
		+ \lambda_{\text{sem}} \, \ell_{\mathrm{sem}}^{(k)} 
		+ \lambda_{\text{unc}} \, \ell_{\mathrm{unc}}^{(k)} 
		\Big),
	\end{split}
\end{equation}
}
where \( \zeta \) is an exponential decay factor that assigns smaller weights to earlier iterations. 
The homography smoothness term \( \ell_{\mathrm{hg}} \) is applied only to the final iteration output.

\subsection{Uncertainty-Guided Bidirectional Flow Fusion}\label{Method-4}

We leverage the predicted uncertainty to guide a bidirectional flow fusion process, enhancing the reliability of the final estimate. Inspired by SMURF~\cite{stone2021smurf}, we employ a lightweight convolutional network to learn a mapping from the backward flow $\mathbf{F}_{t \rightarrow t-1}$ to the forward flow $\mathbf{F}_{t \rightarrow t+1}$. However, our approach fundamentally departs from prior work in its supervisory signal. Instead of relying on heuristic occlusion masks, we train the fusion network exclusively in high-confidence regions identified by our learned uncertainty. 
Specifically, a region is deemed high-confidence only if both its forward and backward uncertainty estimates are below a given threshold $\theta$:
\begin{equation}
	\mathbf{M}_{\text{f}} =
	\mathbb{1}(\boldsymbol{\sigma}^{2}_{t \rightarrow t+1} < \theta), \quad
	\mathbf{M}_{\text{b}} =
	\mathbb{1}(\boldsymbol{\sigma}^{2}_{t \rightarrow t-1} < \theta),
\end{equation}
where $\mathbf{M}_{\text{f}}$ and $\mathbf{M}_{\text{b}}$ are the resulting reliability masks. During inference, this trained network corrects the initial forward flow. 
The final fused flow $\mathbf{F}^{\text{fused}}_{t \rightarrow t+1}$ is computed by adaptively fusing the original estimate with the prediction $\bar{\mathbf{F}}_{t \rightarrow t+1}$, which is derived from $\mathbf{F}_{t \rightarrow t-1}$ via the lightweight convolutional network:

\begin{equation}
	\begin{cases}
		\mathbf{F}^{\text{fused}}_{t \rightarrow t+1} =
		\mathbf{F}_{t \rightarrow t+1} \odot (1 - \mathbf{M}_{\text{fused}}) +
		\bar{\mathbf{F}}_{t \rightarrow t+1} \odot \mathbf{M}_{\text{fused}},
		\\[8pt]
		\mathbf{M}_{\text{fused}} = (1-\mathbf{M}_{\text{f}}) \odot \mathbf{M}_{\text{b}}.
	\end{cases}
\end{equation}

The fusion mask $\mathbf{M}_{\text{fused}}$ activates this correction precisely where the forward flow is uncertain ($\mathbf{M}_{\text{f}}=0$) but the backward flow is confident ($\mathbf{M}_{\text{b}}=1$).

This uncertainty-based fusion strategy fundamentally differentiates our approach from methods like SMURF, which primarily focus on correcting flow within occluded regions (often identified by forward-backward consistency checks~\cite{meister2018unflow}). 
However, true occlusion does not always equate to poor flow estimation, and conversely, non-occluded regions can still yield highly unreliable flow predictions, as illustrated in Fig.~\ref{fig:fusion_mask}. 
By leveraging the reliability masks $\mathbf{M}_{\text{f}}$ and $\mathbf{M}_{\text{b}}$, our method's correction mechanism extends beyond strict occlusion handling. 
Furthermore, this fusion operates as a lightweight refinement on U$^{2}$Flow outputs, achieving multi-frame benefits without the extensive retraining on large-scale datasets required by methods like SMURF~\cite{stone2021smurf}. (See fusion ablation results in Sec.~\ref{E-A-3}.)

\section{Experiments}
\subsection{Implementation Details}
\noindent\textbf{Dataset:}
To ensure a fair comparison, we evaluate our method on the KITTI~\cite{geiger2013vision, menze2015object} 
and Sintel~\cite{sintel2012} datasets, following the training data schedule of prior works~\cite{liu2019selflow, yuan2024unsamflow, sun2025m2flow}.

\noindent\textbf{Training:} 
Our model 
is trained using Adam~\cite{diederik2014adam} ($\beta_1 = 0.9$, $\beta_2 = 0.999$) with a batch size of 4. The training first runs for 100k iterations on raw data with a fixed learning rate of $2\times10^{-4}$, followed by fine-tuning on the original dataset using the OneCycleLR scheduler~\cite{smith2019super} with a maximum learning rate of $2.5\times10^{-4}$ (100k iterations for KITTI and 50k for Sintel). Augmentation regularization (appearance and spatial transformations) is introduced after 50k iterations, while the edge-aware smoothness term $\ell_{\mathrm{sm}}$ is deactivated. The homography smoothness loss $\ell_{\mathrm{hg}}$ and semantic augmentation loss $\ell_{\mathrm{sem}}$ are activated after 50\% of the fine-tuning iterations. The hyperparameters are set to $K=12$, $[\lambda_{\mathrm{hg}}, \lambda_{\mathrm{sm}}, \lambda_{\mathrm{ar}}, \lambda_{\mathrm{sem}}, \lambda_{\mathrm{unc}}, \tau_{\mathrm{hg}}] = [0.1, 55, 0.02, 0.05, 0.005, 2]$, and $\theta=45$ for Sintel and $35$ for KITTI.

For data augmentation, we follow ARFlow~\cite{liuLearningAnalogyReliable2020}, 
applying appearance transformations (brightness, contrast, saturation, hue, Gaussian blur, \textit{etc.}), 
as well as random flipping and swapping. 
All input images are resized to $256 \times 832$ for KITTI and $448 \times 1024$ for Sintel.

\subsection{Benchmark Testing}

We evaluate our method using standard optical flow metrics, including the average endpoint error (EPE) and the percentage of erroneous pixels (Fl). 
Comparisons are conducted against both supervised and unsupervised approaches on the KITTI and Sintel benchmarks. 
As summarized in Tab.~\ref{tab:merged_benchmarks}, our methods, U$^{2}$Flow and U$^{2}$Flow (+FF), achieve highly competitive results, surpassing all existing unsupervised methods on both KITTI-2015 and Sintel benchmarks. 
Specifically, on KITTI-2015, U$^{2}$Flow attains Fl-all=6.13\%, significantly outperforming the previous state-of-the-art unsupervised two-frame method UPFlow~\cite{luoUPFlowUpsamplingPyramid2021} (9.38\%). On KITTI-2012, our method achieves a comparable EPE of 1.4. 
Furthermore, U$^{2}$Flow even surpasses approaches that exploit additional information on KITTI-2015, including the multi-frame methods M2Flow~\cite{sun2025m2flow} (Fl-all=7.37\%) and SMURF~\cite{stone2021smurf} (Fl-all=6.83\%), as well as the semantics-guided methods SemARFlow~\cite{yuan2023semarflow} (Fl-all=8.38\%) and UnSAMFlow~\cite{yuan2024unsamflow} (Fl-all=7.83\%).

The enhanced variant, U$^{2}$Flow (+FF), further improves performance to an Fl-all=6.00\% on KITTI-2015, validating the effectiveness of our bidirectional flow fusion module.

Similarly, on the Sintel benchmark, both U$^{2}$Flow and U$^{2}$Flow (+FF) achieve strong results, consistently outperforming prior unsupervised methods across both the clean and final passes.

\begin{table}[tb]
	\centering
	\fontsize{9pt}{11pt}\selectfont
	\setlength{\tabcolsep}{2.5mm}
	
	\begin{tblr}{
			colspec={l | c c | c c},
			stretch=0.9,
			width=\columnwidth,
		}
		\hline[1.25pt]
		\SetCell[r=2, c=1]{c}{Method} & \SetCell[c=2]{c}{Sintel} & & \SetCell[c=2]{c}{KITTI} & \\
		\cline{2-5}
		& AUSE $\downarrow$ & CC $\uparrow$ & AUSE $\downarrow$ & CC $\uparrow$ \\
		\hline
		FB Check~\cite{meister2018unflow} & 0.19 & 0.57 & 0.21 & 0.53 \\
		PDC-Net+~\cite{truong2023pdc} & 0.18 & 0.45 & 0.16 & 0.50 \\
		Smoothness~\cite{abdein2025self} & 0.26 & 0.43 & 0.23 & 0.45 \\
		U\textsuperscript{2}Flow (Ours) & \textbf{0.11} & \textbf{0.66} & \textbf{0.12} & \textbf{0.64} \\
		\hline[1.25pt]
	\end{tblr}
	\caption{Comparison of uncertainty estimation performance on the Sintel (final, clean) and KITTI (2012, 2015) training sets. }
	\label{tab:ablation_auc_cc}
	
	\vspace{-1.0em}
\end{table}

\begin{figure}[htb]
	\centering
	\includegraphics[width=0.38\textwidth]{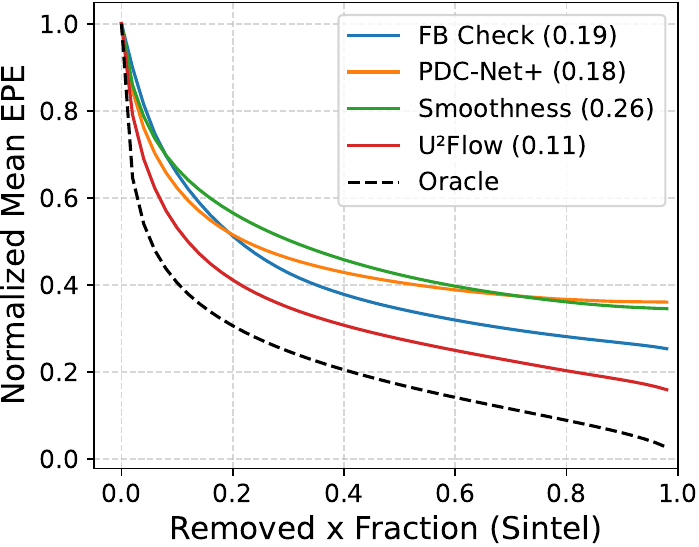}\\[4pt]
	\includegraphics[width=0.38\textwidth]{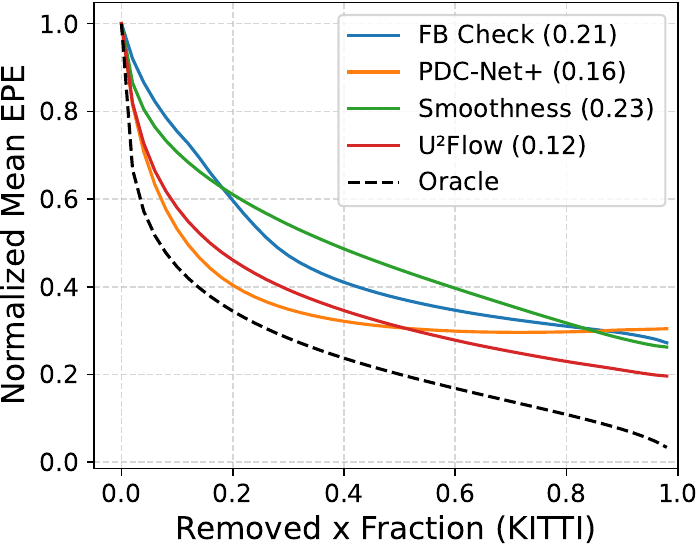}
	\caption{Sparsification curves for uncertainty evaluation. Lower AUSE (shown in parentheses) is better.}
	
	\label{fig:sparsification_curves}
	\vspace{-0.5em}
\end{figure}

\begin{figure*}[htbp]
	\centering
	\includegraphics{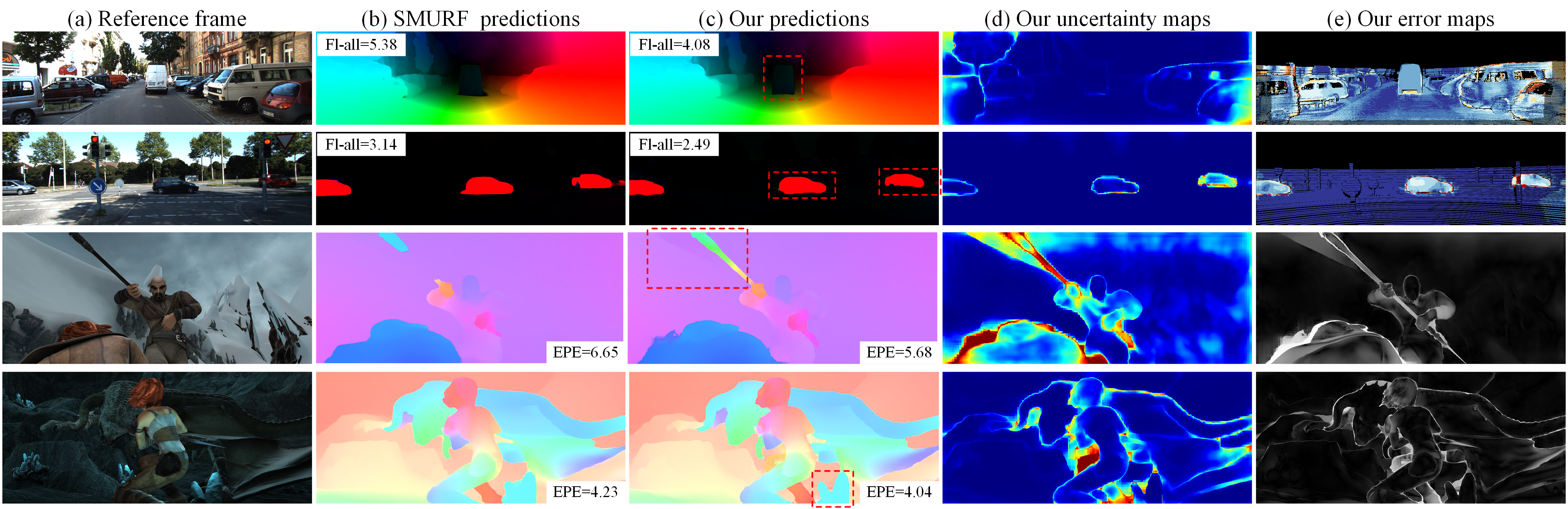}
	\caption{
		Qualitative results on the KITTI test set (sample frames \#5 and \#9) and the Sintel (final pass) test set (samples: \textit{ambush\_3}, frame 23; \textit{cave\_3}, frame 16), compared with SMURF~\cite{stone2021smurf}. Additional examples can be found on the official benchmark website.
	}

	\label{fig:qualitative}
		\vspace{-0.5em}
\end{figure*}

\subsection{Uncertainty Evaluation}

We evaluate the reliability of our uncertainty estimates by measuring their correlation with the ground-truth flow error (EPE)~\cite{ilg2018uncertainty,wannenwetsch2017probflow}. A high-quality uncertainty prediction should correspond strongly to a large estimation error. To this end, we employ two standard quantitative metrics: the Area Under the Sparsification Error curve (AUSE)~\cite{ilg2018uncertainty} and Spearman’s Rank Correlation Coefficient (CC)~\cite{wannenwetsch2017probflow}.

The AUSE evaluates how well predicted uncertainty serves as a proxy for true error. It is derived from a sparsification process that compares the error curve when removing pixels by predicted uncertainty versus by ground-truth error. Consequently, a lower AUSE indicates a better uncertainty estimation. Spearman's CC directly measures the monotonic relationship between predicted uncertainty and true error, with higher values denoting stronger correlation.

Tab.~\ref{tab:ablation_auc_cc} reports uncertainty estimation on the Sintel and KITTI training sets. 
U$^{2}$Flow consistently achieves lower AUSE and higher CC than all baselines, outperforming heuristic methods~\cite{meister2018unflow, abdein2025self} as well as the model-based PDC-Net+~\cite{truong2023pdc}, despite the latter being trained on dense synthetic ground-truth data and fine-tuned on real-world sequences. 

The sparsification curves in Fig.~\ref{fig:sparsification_curves} provide further insight. They show that U$^{2}$Flow produces a more robust and globally consistent ranking of flow errors on both Sintel and KITTI. 
While PDC-Net+ slightly outperforms us in the initial portion on KITTI due to real-data fine-tuning, U$^{2}$Flow’s curve remains closer to the oracle across a wider range of removed fractions, resulting in lower final errors and confirming that our predicted uncertainty reliably reflects overall flow quality.

These results demonstrate that U$^{2}$Flow, guided by augmentation consistency in self-supervised training, effectively captures diverse uncertainty patterns arising from appearance variations and geometric ambiguities.

\subsection{Qualitative Results}
Qualitative samples from the Sintel and KITTI datasets are presented in Fig.~\ref{fig:qualitative}.
Compared with the state-of-the-art competitor~\cite{stone2021smurf}, our method generally achieves superior performance, particularly around motion boundaries. 
Moreover, the predicted uncertainty maps effectively capture and reflect the magnitude of estimation errors.

\subsection{Ablation Study}\label{E-2}
Ablation experiments were performed to assess the contribution of each proposed module, with all models trained under identical conditions except for the ablated components.

\noindent\textbf{Main Components:} \label{E-A-1}
The effectiveness of our key components is validated through the ablation study summarized in Tab.~\ref{tab:ablation_main_cont}. 
Introducing only the Uncertainty Estimation (UE) module without a decoupling strategy (row 2) degrades performance across all metrics. This highlights the challenge of joint optimization, where the uncertainty learning objective can interfere with the primary task of flow estimation.

This issue is resolved by incorporating our Decoupling (Dec.) strategy (row 3). By separating the learning objectives for flow and uncertainty, we observe consistent improvements on both Sintel and KITTI. This confirms that decoupling is crucial for preserving the stability of self-supervised learning while still benefiting from the regularizing effect of uncertainty supervision. Further adding the Flow Refinement (FR) module (row 4) continues this trend, demonstrating the benefit of using uncertainty to guide the network's inference process.

\begin{table}[htb]
	\centering
	\fontsize{9pt}{11pt}\selectfont
	\begin{tblr}{
					colspec={X[c] X[c] X[c] X[c] X[c] | c c | c c}, 
					stretch=0.9, 
					width=\columnwidth,
				}
			\hline[1.25pt]
			\SetCell[r=2, c=1]{c} {UE} & \SetCell[r=2, c=1]{c} {Dec.} & \SetCell[r=2, c=1]{c} {FR} 
			& \SetCell[r=2, c=1]{c} {$\ell_{\mathrm{hg}}$} & \SetCell[r=2, c=1]{c} {FF}
			& \SetCell[c=2]{c}{Sintel} & & \SetCell[c=2]{c}{KITTI 2015} & \\
			\cline{6-9}
			& & & & & Final & Clean & EPE & Fl-all \\
			\hline
			- & - & - & - & - & 2.46 & 1.56 & 1.98 & 6.82 \\
	
			$\checkmark$ & - & - & - & - & 2.57 & 1.65 & 2.18 & 7.72 \\
	
			$\checkmark$ & $\checkmark$ & - & - & - & 2.41 & 1.44 & 1.98 & 6.87 \\
			
			$\checkmark$ & $\checkmark$ & $\checkmark$ & - & - & 2.32 & 1.42 & 1.95 & 6.85 \\
	
			$\checkmark$ & $\checkmark$ & - & $\checkmark$ & - & 2.52 & 1.48 & 1.87 & 6.69 \\
			
			$\checkmark$ & $\checkmark$ & $\checkmark$ & $\bigcirc$ & - & 2.32 & 1.42 & 1.83 & 6.59 \\
			
			\SetRow{bg=gray!20}
			$\checkmark$ & $\checkmark$ & $\checkmark$ & $\bigcirc$ & $\checkmark$ & \textbf{2.29} & \textbf{1.36} & \textbf{1.74} & \textbf{6.30} \\
			\hline[1.25pt]
		\end{tblr}	
	\caption{
			Ablation study on key components. 
			We evaluate the impact of Uncertainty Estimation (UE), Decoupling of flow and uncertainty learning (Dec.), Flow Refinement (FR), uncertainty-enhanced $\ell_{\mathrm{hg}}$, and Flow Fusion (FF). 
			The $\bigcirc$ symbol indicates that $\ell_{\mathrm{hg}}$ is applied only on the KITTI dataset.
		}

	\label{tab:ablation_main_cont}
\end{table}

The uncertainty-enhanced homography loss ($\ell_{\mathrm{hg}}$) offers dataset-specific benefits (rows 5 and 6). On KITTI, where scenes frequently involve substantial planar rigid motion, this provides a significant boost in performance. However, on Sintel, which features complex, non-rigid motion, high uncertainty often correlates with dynamic objects where the homography assumption is fragile. Applying this loss can therefore be detrimental. Consequently, we only apply uncertainty-enhanced $\ell_{\mathrm{hg}}$ to the KITTI dataset, as indicated by the $\bigcirc$ symbol in the table.

\begin{table}[htb]
	\centering
	\fontsize{9pt}{11pt}\selectfont
	\setlength{\tabcolsep}{3mm}
	
	\begin{tblr}{
			colspec={l | c c | c c},
			stretch=0.9,
			width=\columnwidth,
		}
		\hline[1.25pt]
		\SetCell[r=2, c=1]{c}{Model Variant} & \SetCell[c=2]{c}{Sintel} & & \SetCell[c=2]{c}{KITTI 2015} & \\
		\cline{2-5}
		& Final & Clean & EPE & Fl-all \\
		\hline
		w/o Refinement & 2.41 & 1.44 & 1.87 & 6.69 \\
		Refinement w/o Uncertainty & 2.42 & 1.51 & 1.93 & 6.81 \\
		\SetRow{bg=gray!20}
		Refinement w/ Uncertainty & \textbf{2.32} & \textbf{1.42} & \textbf{1.83} & \textbf{6.59} \\
		\hline[1.25pt]
	\end{tblr}	
	\caption{Ablation study on the flow refinement module.
	\label{tab:ablation_flow_module}
}
\vspace{-0.5em}
\end{table}

\begin{table}[htb]
	\centering
	
	\fontsize{9pt}{11pt}\selectfont
	\setlength{\tabcolsep}{2.5mm}
	\resizebox{\columnwidth}{!}{
	\begin{tblr}{
			colspec={Q[l,2.25cm] | Q[c,0.5cm] Q[c,0.6cm] | Q[c,0.78cm] Q[c,0.90cm] Q[c,0.90cm]},
			stretch=0.9,
			width=\columnwidth,
		}
		\hline[1.25pt]
		\SetCell[r=2, c=1]{c}{Method} & \SetCell[c=2]{c}{Sintel} & & \SetCell[c=3]{c}{KITTI 2015} & & \\
		\cline{2-6}
		& Final & Clean & Fl-all & Fl-noc & Fl-occ\\
		\hline
		w/o Fusion & 2.32 & 1.42 & 6.59 & 5.10 & 16.01\\
		Occ Mask& 2.34 & 1.42 & 8.17 & 5.07 & 27.73\\
		\SetRow{bg=gray!20}
		Unc Mask (Ours)& \textbf{2.29} & \textbf{1.36} & \textbf{6.30} & \textbf{5.05} & \textbf{14.17}\\
		\hline[1.25pt]
	\end{tblr}	
	\vspace{-0.5em}
}

	\caption{
		Ablation study on the bidirectional flow fusion module.
	}

	\label{tab:ablation_fusion}
	\vspace{-1.0em}
\end{table}

Finally, our full model (last row) integrates all components including Flow Fusion (FF). It achieves the best overall performance, reaching 2.29/1.36 on Sintel and 1.74/6.30 on KITTI. This result confirms that our complete design can provide uncertainty estimates while delivering optical flow that surpasses the baseline in accuracy.

\noindent\textbf{Flow Refinement Module:} \label{E-A-2}
We conduct an ablation study to evaluate the proposed uncertainty-aware flow refinement mechanism, which integrates estimated uncertainty to dynamically scale flow features and refine flow estimates (Sec.~\ref{Method-2}). Three variants are defined for comparison: (1) Refinement w/ Uncertainty.
(2) w/o Refinement: The model removes the refinement mechanism entirely. The residual is estimated only from the hidden feature, equivalent to standard RAFT-style update: $\Delta \mathbf{F}^{(k)} = \mathcal{C}'_{\text{flow}}(\mathbf{h}^{(k)})$.
(3) Refinement w/o Uncertainty: The model retains the refinement structure but excludes the uncertainty branch. The flow residual is refined using only the unscaled flow feature and a dummy zero-map instead of the uncertainty map: $\Delta \mathbf{F}^{(k)} = \mathcal{C}'_{\text{flow}}(\text{concat}(\mathbf{f}^{(k)}, \mathbf{f}^{(k)}, \mathbf{0}))$. This variant isolates the benefit of feature concatenation alone, removing the dynamic influence of uncertainty.

The ablation results are presented in Tab.~\ref{tab:ablation_flow_module}. Performance metrics on the Sintel and KITTI-2015 training sets clearly demonstrate that the refinement architecture, incorporating dynamic scaling based on uncertainty, is crucial for achieving optimal performance.

\noindent\textbf{Effectiveness of Uncertainty-Guided Fusion:} \label{E-A-3}
The core innovation of our bidirectional flow fusion module is the use of an uncertainty-driven mask, $\mathbf{M}_{\text{fused}}$, to intelligently merge forward and backward flows. To rigorously validate this approach, we compare our full model against two alternatives: one using a traditional occlusion mask for fusion, and another disabling the fusion module entirely.

\begin{table}[htb]
	\centering

	\fontsize{9pt}{11pt}\selectfont
	\setlength{\tabcolsep}{3mm}
	
	\begin{tblr}{
			colspec={l | X[c] X[c] | X[c] X[c]},
			stretch=0.9,
			width=\columnwidth,
		}
		\hline[1.25pt]
		\SetCell[r=2, c=1]{c}{Method} & \SetCell[c=2]{c}{KITTI$\rightarrow$Sintel} & & \SetCell[c=2]{c}{Sintel$\rightarrow$KITTI} & \\
		\cline{2-5}
		& Final & Clean & EPE & Fl-all \\
		\hline
		Ours (w/o Fusion) & 5.80 & 4.67 & 5.10 & 17.05 \\
		Ours (w/ Fusion) & \textbf{5.49} & \textbf{4.23} & \textbf{4.77} & \textbf{16.58} \\
		\hline[1.25pt]
	\end{tblr}	
	
	\caption{Generalization ability. Training on one dataset and testing directly on the other dataset.
	}
	\label{tab:generalization_ability}
	\vspace{-1.5em}
\end{table}

As shown in Tab.~\ref{tab:ablation_fusion}, a particularly notable finding is that guiding fusion with a conventional occlusion mask yields results even worse than the baseline without any fusion. This seemingly counterintuitive outcome arises from a fundamental limitation of binary occlusion flags: they fail to differentiate between high- and low-quality flow estimates within the occluded regions. As a result, valid forward flows are often indiscriminately discarded and replaced with backward-warped counterparts of inferior quality, thereby degrading the overall flow field.

In contrast, our uncertainty-based strategy effectively resolves this ambiguity. By assigning fine-grained, per-pixel confidence scores, it enables more informed fusion decisions that preserve reliable flow estimates and improve the handling of high-uncertainty flow in occluded regions. These results demonstrate that the proposed uncertainty mechanism effectively guides downstream flow fusion and validates the reliability of uncertainty estimates.

\subsection{Generalization Ability}
To evaluate the generalization ability of uncertainty estimation, we test whether the uncertainty predicted by a model trained on one domain can still reliably guide the bidirectional flow fusion task when tested directly on another domain without fine-tuning. As shown in Tab.~\ref{tab:generalization_ability}, the model with our uncertainty-guided fusion module exhibits significantly stronger domain generalization. This indicates that the reliability signal captured by our uncertainty estimation is fundamental and transferable.

\section{Conclusion}
We propose U$^{2}$Flow, a recurrent unsupervised framework for jointly estimating optical flow and per-pixel uncertainty. Our method leverages augmentation consistency and a decoupled learning strategy to achieve stable training of both flow and uncertainty in unsupervised settings. We demonstrate that the learned uncertainty is not merely a byproduct, but a valuable signal that can effectively guide adaptive flow refinement, modulate smoothness constraints, and enable robust bidirectional flow fusion. 
U$^{2}$Flow establishes a state-of-the-art for unsupervised optical flow while producing highly reliable uncertainty estimates.

\vspace{1em}
\noindent\textbf{Limitations}~~
Our uncertainty supervision relies on the model's predictive variance under a predefined set of augmentations, which may not fully capture all real-world sources of error, such as severe non-Gaussian motion blur or atmospheric distortions. In future work, we plan to enrich the augmentation space using generative models to synthesize more realistic and diverse image degradations.

\newpage

\section*{Acknowledgments}
This research was supported by the Guangzhou Basic and Applied Basic Research Foundation under Grant
SL2024A04J0183, the Guangxi Key Research and Development Project under Grant GuikeAB25069495, the National Natural Science Foundation of China under Grant 92470202, and the Fund of National Key Laboratory of Multispectral Information Intelligent Processing Technology (No. 202410487201).

{
    \small
    \bibliographystyle{ieeenat_fullname}
    \bibliography{main}
}


\clearpage
\setcounter{page}{1}
\maketitlesupplementary
\appendix

\section{Method details}
\subsection{Recurrent update block}
\begin{figure*}[htb]
	\centering
	\includegraphics[width=\textwidth]{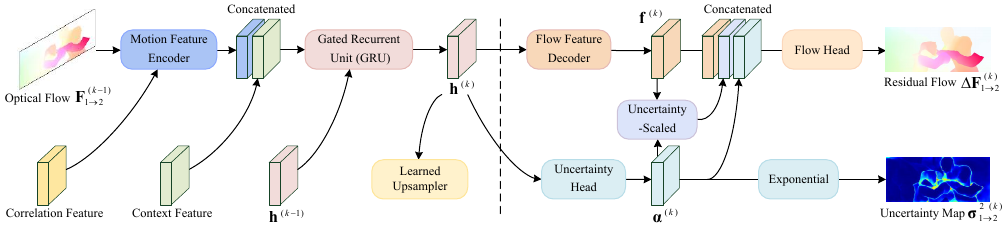}
	\caption{Detailed illustration of the recurrent update block. The diagram depicts the process flow for the $k$-th iteration. Components to the left of the dashed line represent the original RAFT~\cite{teedRAFTRecurrentAllPairs2021} architecture, while components to the right constitute our proposed flow refinement module and uncertainty estimation head.}
	\label{fig:appendix_update}
	\vspace{-1.0em}
\end{figure*}

Fig.~\ref{fig:appendix_update} provides a detailed illustration of our recurrent update block, highlighting the processing flow within a single iteration \(k\).

At each iteration, the block takes three inputs: the previous optical flow estimate \( \mathbf{F}^{(k-1)}_{1 \rightarrow 2} \), the context feature extracted from the reference image, and the correlation feature retrieved from the 4D correlation volume using the previous flow estimate. The flow estimate and correlation feature are fused into a motion feature, which is then concatenated with the context feature. This combined tensor is fed into a Gated Recurrent Unit (GRU), which updates its hidden state to \( \mathbf{h}^{(k)} \), with the initial state \( \mathbf{h}^{(0)} \) initialized from the context feature.

From the updated hidden state, our prediction heads generate two outputs: a residual flow update \( \Delta \mathbf{F}^{(k)}_{1 \rightarrow 2} \) and a corresponding per-pixel uncertainty map \( \boldsymbol{\sigma}^{2\,(k)}_{1 \rightarrow 2} \). Finally, a learned upsampling module upsamples the refined flow and uncertainty to the full resolution of the input image.

\begin{figure*}[htb]
	\centering
	\includegraphics[width=\textwidth]{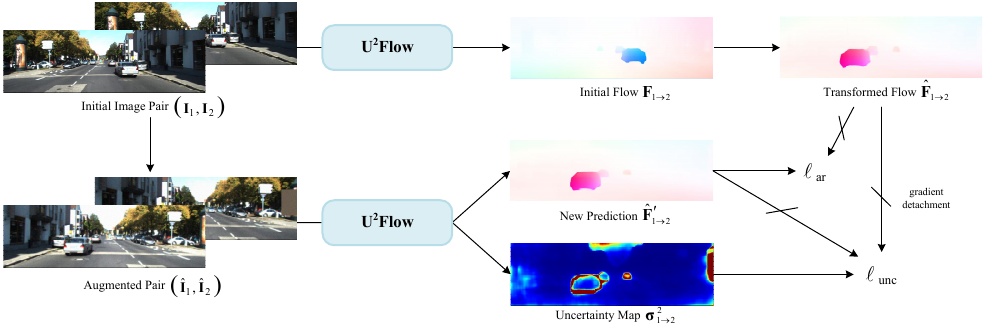}
	\caption{Schematic diagram of the self-supervision process. (1) An initial flow $\mathbf{F}_{1 \rightarrow 2}$ is predicted from the original image pair. (2) Strong augmentations are applied to both the images and the flow, creating an augmented image pair and a transformed pseudo-ground-truth flow $\hat{\mathbf{F}}_{1 \rightarrow 2}$. (3) A new flow prediction $\hat{\mathbf{F}}^{\prime}_{1 \rightarrow 2}$ is generated from the augmented images. (4) The inconsistency $\hat{D}$ between the pseudo-ground-truth and the new prediction serves as the target signal for both the augmentation loss ($\ell_{\mathrm{ar}}$) and the uncertainty supervision loss ($\ell_{\mathrm{unc}}$).}
	\label{fig:appendix_loss}
	\vspace{-1.0em}
	
\end{figure*}

\begin{table*}[tb]
	\centering
	\fontsize{9pt}{11pt}\selectfont
	\setlength{\tabcolsep}{2.5mm}
	
	\begin{tblr}{
			colspec={c | c c c| c c c c | c},
			stretch=0.9,
			width=\columnwidth,
		}
		\hline[1.25pt]
		\SetCell[r=2, c=1]{c}{Iter. Num.} & \SetCell[c=3]{c}{Sintel Final} & & & & \SetCell[c=5]{c}{KITTI 2015} & & & &\\
		\cline{2-10}
		& EPE $\downarrow$ & AUSE $\downarrow$ & CC $\uparrow$ & EPE $\downarrow$ & Fl-all $\downarrow$ & AUSE $\downarrow$ & CC $\uparrow$ & Runtime\\
		\hline
		2 & 3.52 & 0.11 & 0.69 & 3.71 & 12.31 & 0.20 & 0.53 & 18 ms\\
		4 & 2.75 & 0.10 & 0.69 & 2.32 & 8.23 & 0.21 & 0.50 & 35 ms\\
		8 & 2.41 & 0.11 & 0.68 & 1.90 & 6.75 & 0.22 & 0.48 & 52 ms\\
		12 & 2.32 & 0.11 & 0.67 & 1.83 & 6.59 & 0.22 & 0.48 & 66 ms\\
		16 & 2.30 & 0.11 & 0.67 & 1.82 & 6.53 & 0.22 & 0.47 & 97 ms\\
		\hline[1.25pt]
	\end{tblr}
	\caption{Ablation study on the number of recurrent iterations ($K$). Flow accuracy improves and saturates with more iterations, while uncertainty estimation (AUSE) remains robustly stable. Runtime is measured on KITTI-2015.}
	\label{tab:appendix_num_iter}
	
\end{table*}

\subsection{Augmentation and uncertainty supervision}
Fig.~\ref{fig:appendix_loss} provides a schematic overview of our self-supervision strategy, which generates the signals for both the augmentation loss ($\ell_{\mathrm{ar}}$) and the uncertainty supervision loss ($\ell_{\mathrm{unc}}$).

Specifically, the process begins by computing an initial flow estimate, $\mathbf{F}_{1 \rightarrow 2}$, for an image pair $(\mathbf{I}_{1}, \mathbf{I}_{2})$ in a forward pass. Subsequently, we apply a set of strong appearance augmentations (e.g., color jitter, contrast adjustment, Gaussian noise, random erase) and spatial transformations (e.g., translation, rotation, rescaling) to both the images and the flow field. This produces an augmented pair $(\hat{\mathbf{I}}_{1}, \hat{\mathbf{I}}_{2})$ and a transformed pseudo-ground-truth flow $\hat{\mathbf{F}}_{1 \rightarrow 2}$. The network then re-estimates the flow for the augmented pair, yielding a new prediction $\hat{\mathbf{F}}^{\prime(k)}_{1 \rightarrow 2}$ at iteration $k$.

The per-pixel $\ell_{1}$ distance between these two flow fields, denoted as $\hat{D}^{(k)}(\boldsymbol{p}) = \|\hat{\mathbf{F}}_{1 \rightarrow 2}(\boldsymbol{p}) - \hat{\mathbf{F}}^{\prime(k)}_{1 \rightarrow 2}(\boldsymbol{p})\|_1$, serves as the self-supervised target, as it directly captures the model's predictive inconsistency under perturbation. This target inconsistency, $\hat{D}^{(k)}$, is then leveraged in two distinct ways, as detailed in Eq.~9 and Eq.~10  in the main paper:
\begin{itemize}
	
	\item \textbf{For the uncertainty loss ($\ell_{\mathrm{unc}}$):} The inconsistency serves as the supervisory signal within our maximum likelihood objective to train the uncertainty head.
	\item \textbf{For the augmentation loss ($\ell_{\mathrm{ar}}$):} We directly minimize this inconsistency to enforce model robustness against perturbations.

\end{itemize}

In addition to these augmentations, and as described in the main paper, we also incorporate a semantic augmentation loss, $\ell_{\mathrm{sem}}$~\cite{yuan2024unsamflow}. This loss follows a similar formulation to $\ell_{\mathrm{ar}}$ but is computed on semantically augmented data, where object regions are randomly copied and pasted between image pairs. This process introduces more realistic occlusions and compositional variations, further enhancing model robustness.

\section{Result details}
\subsection{Implementation details}
Our model is tested on Ubuntu 18.04 with Python 3.7.16, PyTorch 1.13.1, Torchvision 0.14.1, and CUDA 11.3. We use 2 NVIDIA RTX 3090 GPUs with 24 GB of memory each. More details can be found in the code appendix.

\subsection{Efficiency analysis}
We evaluate the computational efficiency of our network. For each RGB sample of resolution $376\times1242$, our model achieves an average inference time of 0.0663 seconds ($\sim$15FPS) on an NVIDIA RTX 3090 GPU. 
In terms of model size, our network contains 5.22M parameters, making it marginally more compact than the original RAFT~\cite{teedRAFTRecurrentAllPairs2021} architecture (5.26M). This parameter efficiency is attributed to the lightweight design of our proposed flow refinement module and uncertainty estimation head.

\subsection{Ablation study on the number of recurrent iterations}

To analyze the impact of the recurrent refinement process, we performed an ablation study on the number of iterations ($K$) using the Sintel (final pass) and KITTI-2015 training sets. The results are summarized in Table \ref{tab:appendix_num_iter}.

As expected, the accuracy of the optical flow estimates generally improves with an increasing number of iterations across both datasets. The performance gains exhibit diminishing returns and begin to saturate at approximately $K=12$ iterations. This trend confirms the effectiveness of the iterative refinement process while justifying our choice of $K=12$ for the final model to balance performance and efficiency.

Interestingly, the quality of our uncertainty estimation, measured by AUSE, remains stable across different values of $K$. This suggests that the uncertainty head relies on intrinsic feature properties rather than the final converged flow, demonstrating the robustness of our uncertainty learning mechanism.

\begin{table}[htp]
	\centering
	\fontsize{9pt}{11pt}\selectfont
	\setlength{\tabcolsep}{2.5mm}
	
	\begin{tblr}{
			colspec={l | c c | c c},
			stretch=0.5,
			width=\columnwidth,
		}
		\hline[1.25pt]
		\SetCell[r=2, c=1]{c}{Method} & \SetCell[c=2]{c}{Sintel} & & \SetCell[c=2]{c}{KITTI} & \\
		\cline{2-5}
		& AUSE $\downarrow$ & CC $\uparrow$ & AUSE $\downarrow$ & CC $\uparrow$ \\
		\hline
		Appearance Aug. Only & 0.46 & 0.25 & 0.25 & 0.51 \\
		Spatial Aug. Only & 0.12 & 0.65 & 0.12 & 0.64 \\
		Full (Ours) & \textbf{0.11} & \textbf{0.66} & \textbf{0.12} & \textbf{0.64} \\ 
		\hline[1.25pt]
	\end{tblr}
	\caption{Ablation study of augmentation strategies for uncertainty.
	}
	\label{tab:ablation_aug_unc}
\end{table}

\begin{figure*}[htb]
	\centering
	\includegraphics[width=0.9\textwidth]{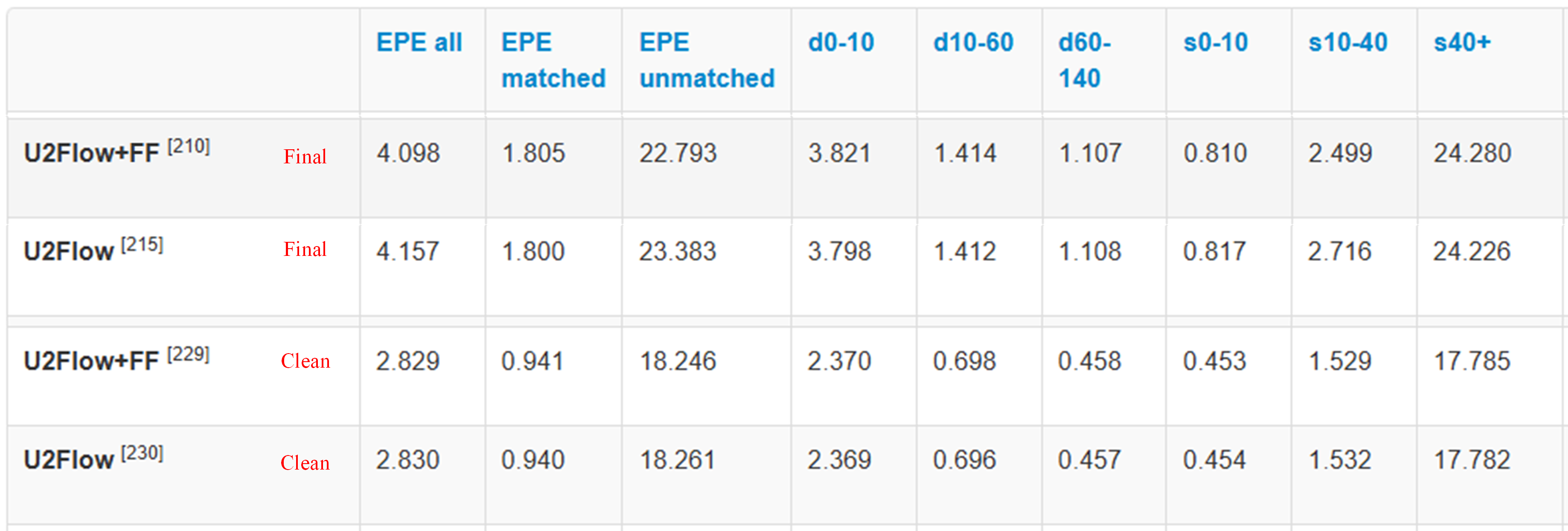}
	\caption{Detailed test results of our model on Sintel~\cite{sintel2012}.}
	\label{fig:appendix_benchmark_sintel}
\end{figure*}

\begin{figure}[htbp]
	\centering
	\resizebox{\columnwidth}{!}{
		\subfloat[KITTI-2015 results
		\label{fig:subfig2015}]{
			\includegraphics{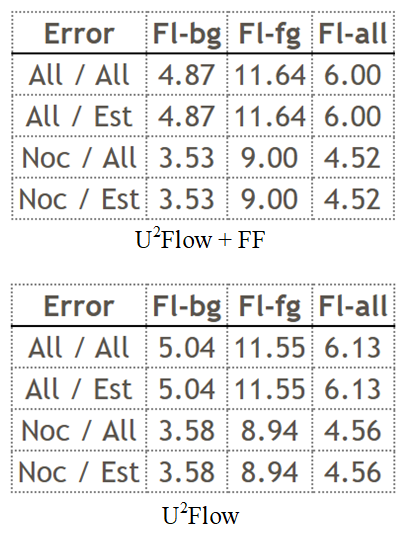}
		}
		
		\subfloat[KITTI-2012 results
		\label{fig:subfig2012}]{
			\includegraphics{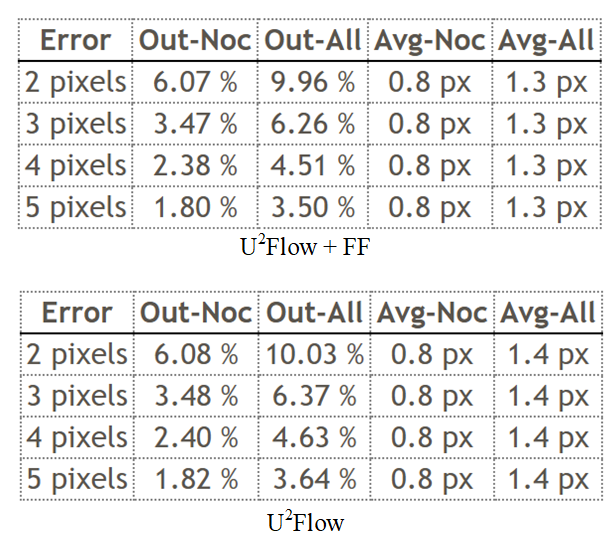}
		}
	}
	\vspace{-0.5em}
	\caption{Detailed test results of our final model on KITTI~\cite{geiger2013vision, menze2015object}.}
	\label{fig:appendix_benchmark_kitti}
	
\end{figure}

\subsection{Ablation study on augmentation sensitivity}
To further analyze the impact of different data augmentation strategies on our method, we provide an ablation study on augmentation types in Tab.~\ref{tab:ablation_aug_unc}. We observe that spatial augmentation is the dominant factor for reliable uncertainty learning: 
using \emph{Spatial Only} achieves performance comparable to \emph{Full} (e.g., AUSE 0.12 vs. 0.11 on Sintel), while \emph{Appearance Only} degrades significantly (AUSE 0.46). 
Appearance augmentation offers negligible gain on KITTI and a minor improvement on Sintel, likely due to its synthetic lighting. 

This is consistent with optical flow being primarily a geometric correspondence task, where spatial perturbations provide stronger uncertainty supervision than photometric variations.

\begin{table}[htb]
	\vspace{-1.2em}
	\centering
	\fontsize{9pt}{11pt}\selectfont
	\setlength{\tabcolsep}{2.5mm}
	
	\begin{tblr}{
			colspec={l | c c | c c},
			stretch=0.5,
			width=\columnwidth,
		}
		\hline[1.25pt]
		\SetCell[r=2, c=1]{c}{Model Variant} & \SetCell[c=2]{c}{Sintel} & & \SetCell[c=2]{c}{KITTI 2015} & \\
		\cline{2-5}
		& Final & Clean & EPE & Fl-all \\
		\hline
		w/o Uncertainty-Scaled & 2.40 & 1.52 & 2.04 & 7.24 \\
		w/ Uncertainty-Scaled (Ours)& \textbf{2.32} & \textbf{1.42} & \textbf{1.83} & \textbf{6.59} \\
		\hline[1.25pt]
	\end{tblr}	
	\caption{Ablation study on the uncertainty scaling mechanism. Explicitly modulating flow representations with the scaled feature $\tilde{\mathbf{f}}^{(k)}$ yields better performance than implicit direct concatenation.
	}
	\label{tab:ablation_scaling}
	\vspace{-1.3em}
\end{table}

\subsection{Ablation study on uncertainty scaling}
In Eq.~4 of the main paper, we introduce an uncertainty-scaled flow feature $\tilde{\mathbf{f}}^{(k)} = \mathbf{f}^{(k)} \odot \mathbf{s}^{(k)*}$ to modulate flow representations according to the predicted reliability. This operation suppresses unreliable features before propagation, enabling the refinement module to focus on more reliable regions. 
To evaluate the effectiveness of this design, we compare two variants in Eq.~5:

\begin{itemize}
	\item \textbf{w/o Uncertainty-Scaled}: directly concatenate the flow feature 
	$\mathbf{f}^{(k)}$ with the uncertainty prediction $\boldsymbol{\alpha}^{(k)}$, 
	leaving the network to implicitly learn how to utilize the uncertainty information.
	
	\item \textbf{With Uncertainty-Scaled (ours)}: use the uncertainty-scaled feature 
	$\tilde{\mathbf{f}}^{(k)} = \mathbf{f}^{(k)} \odot \mathbf{s}^{(k)*}$ to explicitly 
	modulate the flow representation before feature fusion.
\end{itemize}

As shown in Tab.~\ref{tab:ablation_scaling}, the proposed uncertainty scaling consistently improves performance, indicating that explicitly modulating flow features with predicted uncertainty is more effective than relying on implicit feature fusion.

\begin{figure*}[htb]
	\centering
	\includegraphics[width=\textwidth]{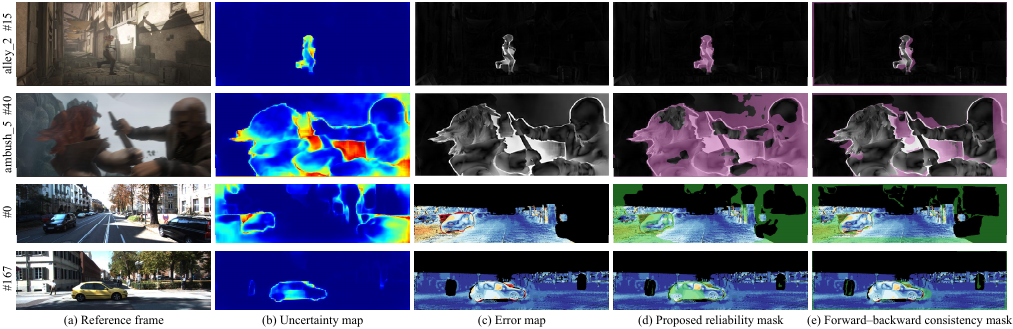}
	\caption{Visualization of selected regions for homography smoothness on Sintel and KITTI. The top two rows show samples from the Sintel dataset, while the bottom two rows are from KITTI. The uncertainty-based reliability mask successfully identifies regions with high flow error in both datasets. However, for Sintel, these high-error regions are typically non-rigid characters for which the planar homography assumption does not hold. In contrast, for KITTI, high-error regions often correspond to rigid vehicles and planar road surfaces, which are well-suited for homography regularization. This illustrates why the uncertainty-enhanced $\ell_{\mathrm{hg}}$ is effective on KITTI but not on Sintel.}
	\label{fig:appendix_hm_loss}
\end{figure*}

\subsection{Analysis of the uncertainty-enhanced homography smoothness loss}

The homography smoothness loss~\cite{yuan2024unsamflow} regularizes flow estimation by enforcing consistency with a planar motion model. In our framework, we employ an uncertainty-based reliability mask to select the regions where this loss is applied. In this section, we further elaborate on the dataset-dependent performance of this uncertainty-enhanced homography smoothness loss ($\ell_{\mathrm{hg}}$), which, as shown in the main paper, improves results on KITTI but degrades them on Sintel.

Our investigation reveals that the core issue is not the quality of the uncertainty mask itself, but rather the nature of the high-uncertainty regions within each dataset. As illustrated in Fig.~\ref{fig:appendix_hm_loss}, the uncertainty-based reliability mask effectively identifies regions with high estimation error on Sintel. 
However, in the Sintel dataset, these high-uncertainty regions frequently correspond to characters and creatures undergoing complex, non-rigid, and non-planar motion (e.g., walking, running, or fighting). 
Applying the homography smoothness loss to these areas forces the network to regularize the flow field based on a planar motion assumption, which is geometrically invalid for such movements. 
This fundamental mismatch introduces erroneous constraints, ultimately degrading the overall flow estimation quality rather than refining it.

In contrast, the forward--backward consistency check~\cite{meister2018unflow}, which is mainly used as an occlusion detector, tends to highlight static or slow-moving background regions that become occluded. As shown in Fig.~\ref{fig:appendix_hm_loss}, these background areas (e.g., walls, ground) are typically planar and thus suitable for homography regularization.

The situation in the KITTI dataset is markedly different. The scenes are dominated by rigid objects (vehicles) and large planar surfaces (roads, buildings). 
Even the primary moving objects like cars are composed of multiple planar surfaces. 
Consequently, the planar motion assumption holds true for many high-uncertainty regions, making the uncertainty-enhanced $\ell_{\mathrm{hg}}$ a highly effective regularizer that provides a significant performance boost.

\subsection{Benchmark test screenshots}
The results of U$^{2}$Flow and U$^{2}$Flow (+FF) have been submitted to the KITTI and Sintel online benchmarks and can be found on the leaderboards by searching for ``U$^{2}$Flow". Screenshots of the detailed evaluation metrics from the official websites are provided in Fig.~\ref{fig:appendix_benchmark_sintel} and Fig.~\ref{fig:appendix_benchmark_kitti}.

\begin{figure*}[htbp]
	\centering
	\includegraphics[width=\linewidth]{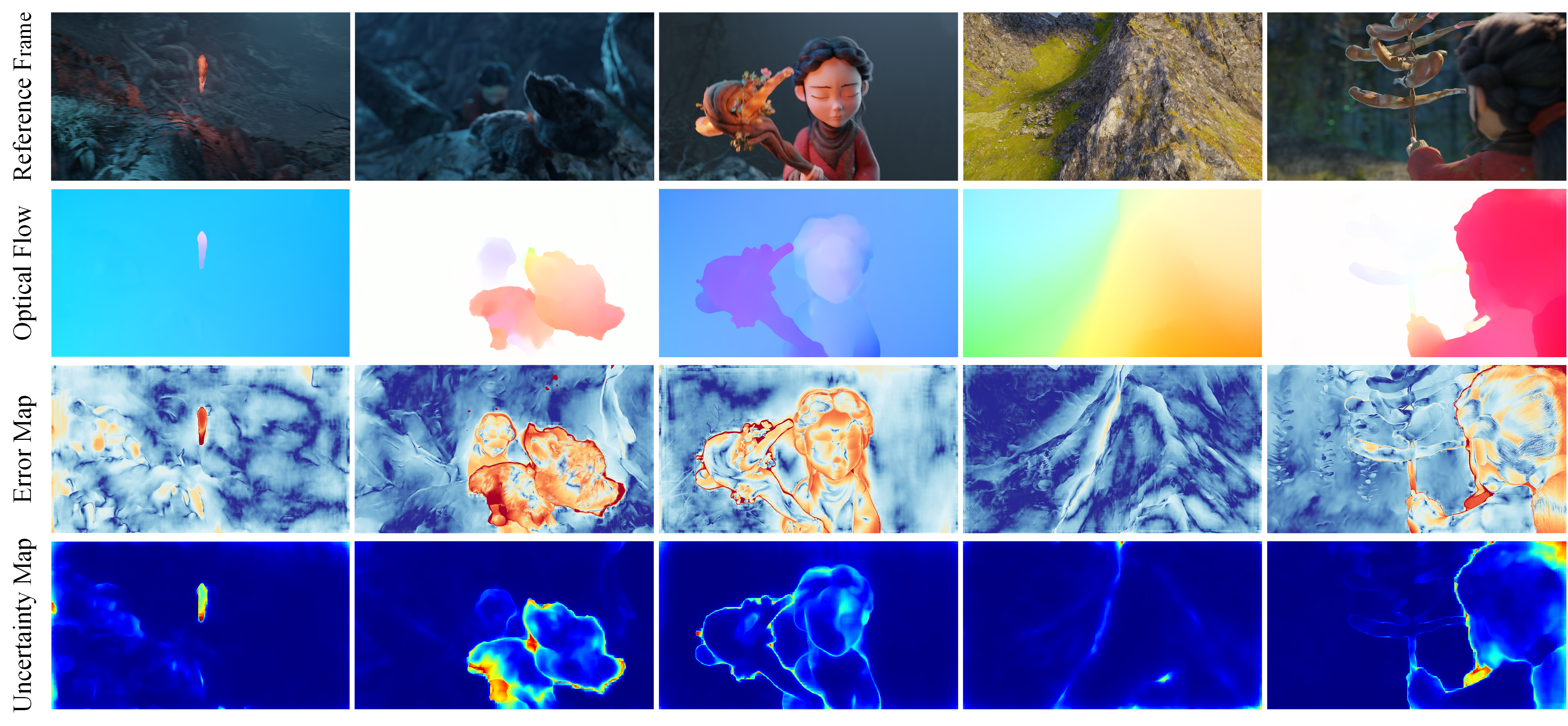}
	\caption{Qualitative results on the Spring benchmark~\cite{mehl2023spring} without fine-tuning.}
	\label{fig:spring}
\end{figure*}

\subsection{Results on the Spring Dataset}

To further demonstrate the generalization capability of our approach, 
we evaluate our model on the Spring~\cite{mehl2023spring} test set without fine-tuning. 
As shown in Tab.~\ref{tab:spring_results}, our U$^2$Flow, as an unsupervised method, consistently outperforms 
SMURF~\cite{stone2021smurf} across most metrics (below the dashed line). 
Moreover, compared with supervised approaches (above the dashed line), which are trained in a multi-stage 
manner on multiple datasets, our model—trained exclusively on Sintel~\cite{sintel2012}—still demonstrates strong generalization ability and surpasses them on several key metrics.

We also provide qualitative results on the Spring dataset in Fig.~\ref{fig:spring}. 
As illustrated, the predicted flow fields are spatially coherent and preserve sharp motion boundaries. 
In addition, the predicted uncertainty maps show a reasonable correlation with the estimation errors, 
suggesting that the model can capture the reliability of its predictions.

\begin{table*}[tb]
	\centering
	\fontsize{9pt}{11pt}\selectfont
	\setlength{\tabcolsep}{2.0mm}
	\resizebox{\linewidth}{!}{
		
	\begin{tblr}{
			colspec={ l | c c c c c c c c c c c c | c c c},
			width=\linewidth,
			stretch=0.9,
			column{14,15,16} = {bg=lightblue}
		}
		\hline[1.25pt]
		\SetCell[r=2]{c}{Model} & \SetCell[c=11]{c}{1px} & & & & & & & & & & & & \SetCell[r=2]{c}{EPE} & \SetCell[r=2]{c}{Fl} & \SetCell[r=2]{c}{WAUC}\\
		\cline{2-13}
		& \SetCell{bg=lightblue} total & low-det. & high-det. & matched & unmat. & rigid & non-rig. & not sky & sky & s0-10 & s10-40 & s40+ & & &\\
		\hline
		RAFT~\cite{teedRAFTRecurrentAllPairs2021} & \SetCell{bg=lightblue} 6.79 & 6.43 & \textbf{64.09} & 6.00 & \underline{39.48} & 4.11 & \textbf{27.09} & \textbf{5.25} & 30.18 & \textbf{3.13} & \textbf{5.30} & 41.40 & 1.476 & 3.20 & \underline{90.92}\\
		GMA~\cite{jiang2021learning} & \SetCell{bg=lightblue} 7.07 & 6.70 & 66.20 & 6.28 & 39.89 & 4.28 & 28.25 & 5.61 & 29.26 & 3.65 & \underline{5.39} & 40.33 & 0.914 & 3.08 & 90.72\\
		GMFlow~\cite{xu2022gmflow} & \SetCell{bg=lightblue} 10.36 & 9.93 & 76.61 & 9.06 & 63.95 & 6.80 & 37.26 & 8.95 & 31.68 & 5.41 & 9.90 & 52.94 & 0.945 & 2.95 & 82.34\\
		FlowFormer~\cite{huang2022flowformer} & \SetCell{bg=lightblue} 6.51 & 6.14 & \underline{64.22} & 5.77 & \textbf{37.29} & \textbf{3.53} & 29.08 & 5.50 & 21.86 & \underline{3.38} & 5.53 & 35.34 & 0.723 & 2.38 & \textbf{91.68}\\
		\hline[dashed]
		SMURF~\cite{stone2021smurf} & \SetCell{bg=lightblue} \underline{6.43} & \underline{6.04} & 66.52 & \underline{5.49} & 45.12 & 3.58 & 27.95 & 5.53 & \underline{20.07} & 3.44 & 6.59 & \underline{30.99} & \underline{0.659} & \underline{2.10} & 90.45 \\
		\textbf{U$^2$Flow (Ours)} & \SetCell{bg=lightblue} \textbf{6.32} & \textbf{5.94} & 66.02 & \textbf{5.38} & 45.31 & \underline{3.56} & \underline{27.26} & \underline{5.46} & \textbf{19.38} & 3.41 & 6.54 & \textbf{30.13} & \textbf{0.608} & \textbf{1.88} & 89.32\\
		\hline[1.25pt]
	\end{tblr}
}
	\caption{Optical flow generalization results on the Spring benchmark~\cite{mehl2023spring} without fine-tuning. 
		We report the 1px outlier rate across low/high-detail, (un)matched, (non-)rigid, and (non-)sky regions, 
		along with the EPE, Fl error, and WAUC~\cite{richter2017playing} metrics. 
		The best and second-best results are highlighted in bold and underline, respectively, 
		while key metrics are emphasized in blue.}
	\label{tab:spring_results}
	\vspace{-1.0em}
\end{table*}

\subsection{More qualitative examples}
\begin{figure*}[htb]
	\centering
	\includegraphics[width=\textwidth]{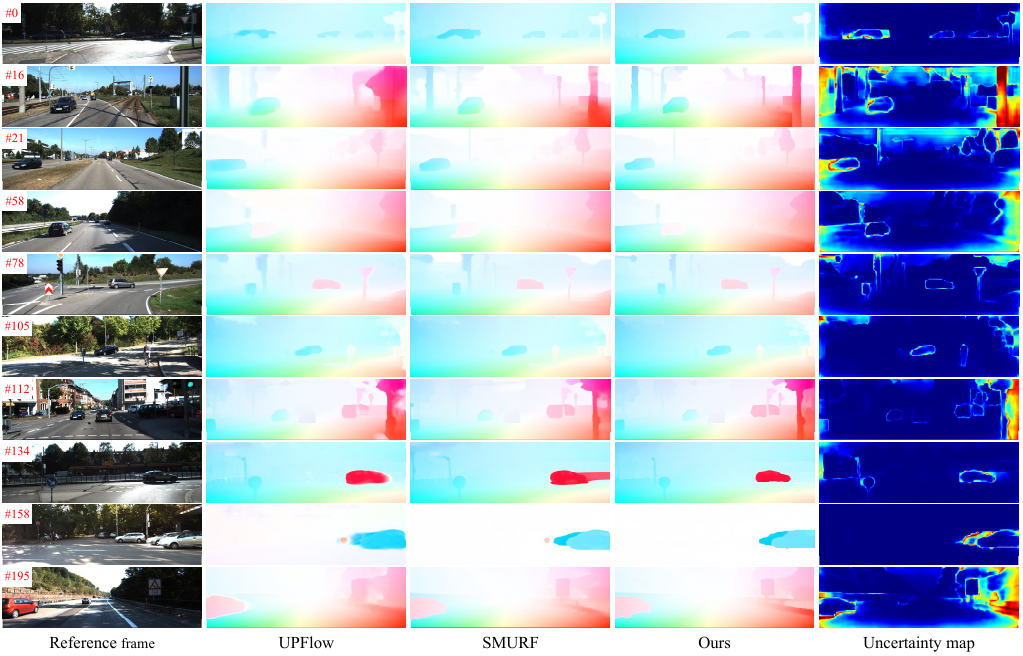}
	\caption{Additional qualitative results on the KITTI-2015 test set.}
	\label{fig:appendix_qualitative_kitti}
\end{figure*}

\begin{figure*}[htb]
	\centering
	\includegraphics[width=\textwidth]{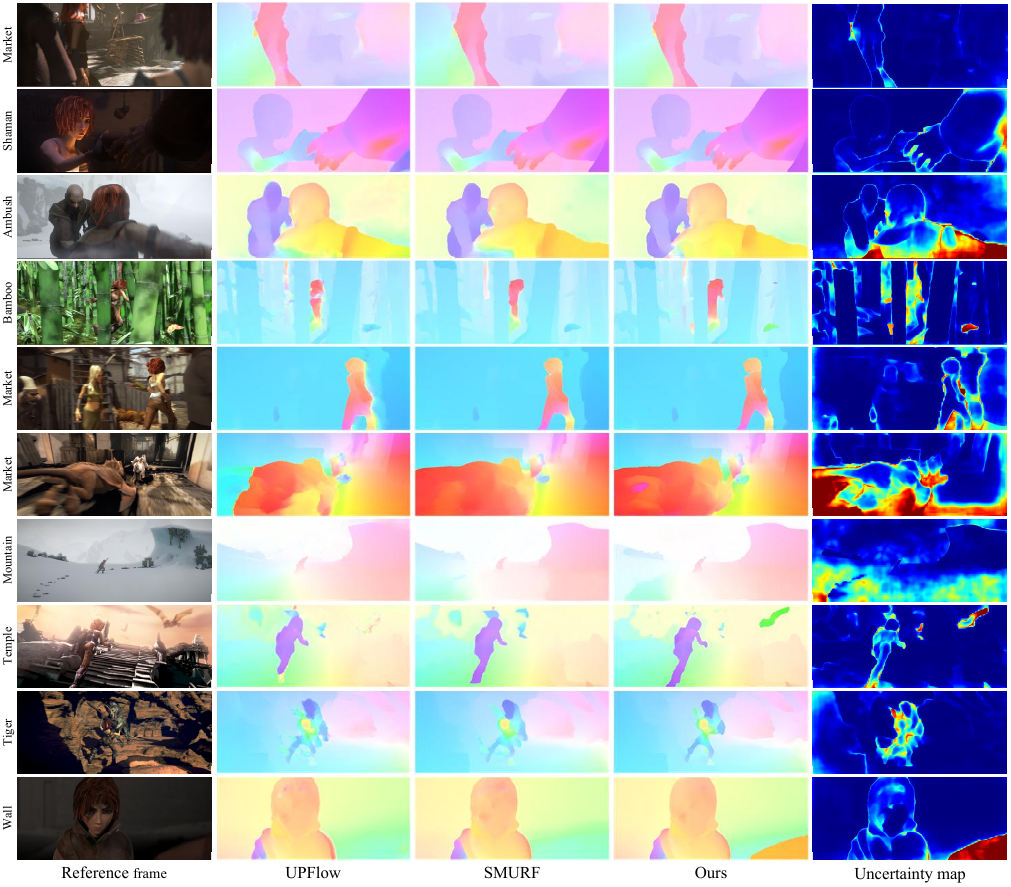}
	\caption{Additional qualitative results on the Sintel (final pass) test set.}
	\label{fig:appendix_qualitative_sintel}
\end{figure*}

We present additional qualitative results from the KITTI-2015 test set (Fig.~\ref{fig:appendix_qualitative_kitti}) and the Sintel test set (Fig.~\ref{fig:appendix_qualitative_sintel}). We compare our method with the state-of-the-art unsupervised two-frame approach UPFlow~\cite{luoUPFlowUpsamplingPyramid2021} and the multi-frame trained model SMURF~\cite{stone2021smurf}, which is also based on RAFT~\cite{teedRAFTRecurrentAllPairs2021}.

\end{document}